\documentclass[runningheads,a4paper]{llncs}

\def\etal{\textit{et al.}\xspace}

\def\cf{\textit{c.f.}~}
\def\wrt{{w.r.t.}~}

\def\transf{{\cal T}}
\def\moving{{I_m}}
\def\warped{{I'_m}}
\def\fixed{{I_f}}
\def\metric{{M}}
\def\domain{{\mathrm{\Omega}}}
\def\regularization{{R}}
\def\param{{\mathbf{\theta}}}

\usepackage{xspace}
\usepackage{amssymb}
\usepackage{graphicx}
\usepackage{amsmath}
\usepackage[ruled]{algorithm2e}
\usepackage{hyperref}

\usepackage{color}

\usepackage{url}
\urldef{\mailsa}\path|{martin.simonovsky,nikos.komodakis}@enpc.fr|
\urldef{\mailsb}\path|gutierrez.becker@tum.de,{mateus,navab}@in.tum.de|

\begin{document}

\mainmatter  

%
\title{A Deep Metric for Multimodal Registration}


\titlerunning{A Deep Metric for Multimodal Registration}

\author{Martin~Simonovsky\inst{1} \and Benjam\'in~Guti\'errez-Becker\inst{2}
Diana~Mateus\inst{2} \and Nassir~Navab\inst{2} \and Nikos~Komodakis\inst{1}}
\authorrunning{M. Simonovsky et al.} 

\institute{Imagine, Universit\'{e} Paris Est / \'{E}cole des Ponts ParisTech, France \\ \mailsa
           \and
           Computer Aided Medical Procedures, Technische Universit{\"a}t M{\"u}nchen, Germany \\ \mailsb}

\toctitle{Lecture Notes in Computer Science}
\tocauthor{Authors' Instructions}
\maketitle

\begin{abstract}
Multimodal registration is a challenging problem in medical imaging due the high variability of tissue appearance under different imaging modalities. The crucial component here is the choice of the right similarity measure. We make a step towards a general learning-based solution that can be adapted to specific situations and present a  metric based on a convolutional neural network. Our network can be trained from scratch even from a few aligned image pairs. The metric is validated on intersubject deformable registration on a dataset different from the one used for training, demonstrating good generalization. In this task, we outperform mutual information by a significant margin.
\end{abstract}

\section{Introduction}
Multimodal registration is a very challenging problem in medical imaging commonly faced during image-guided interventions and data fusion~\cite{sotiras:survey}. The main difficulty of the multimodal registration task comes from the great variability of tissue or organ appearance when imaged by different physical principles, which translates in the lack of a general rule to compare such images. Therefore, efforts to tackle this problem focus mainly on the design of multimodal similarity metrics.

Recent works have explored the use of supervised methods to learn similarity metrics from a set of aligned examples \cite{cao:analogies,lee:learning,michel:boost}, showing potential to outperform hand-crafted metrics in particular applications.
However, a general method to learn similarity between any two modalities calls for higher capacity models.

Inspired by their success in computer vision, we propose to learn such general similarity metric based on Convolutional Neural Networks (CNNs). The problem is modelled as a classification task, where the goal is to discriminate between aligned and misaligned patches from different modalities.
To the best of our knowledge, this is the first time that CNNs are used in the context of multimodal medical image registration.

The ability of our metric to obtain reliable registrations is demonstrated on the ALBERTs database of neonatal images \cite{Gousias:alberts1}, where we outperform Mutual Information \cite{mattes:mi}. Importantly, we train on a separate dataset (IXI database of adults \cite{ixi:ixi}), demonstrating the capability to generalize to data acquired with different scanners and with demographic differences in the subjects. We also show that our method is able to learn reliable multimodal similarities even with a small training set, as is often the case in medical imaging applications.

\begin{figure*}[t]
\centering
\includegraphics[trim={0 12.5cm 0 0},clip,width=1\linewidth]{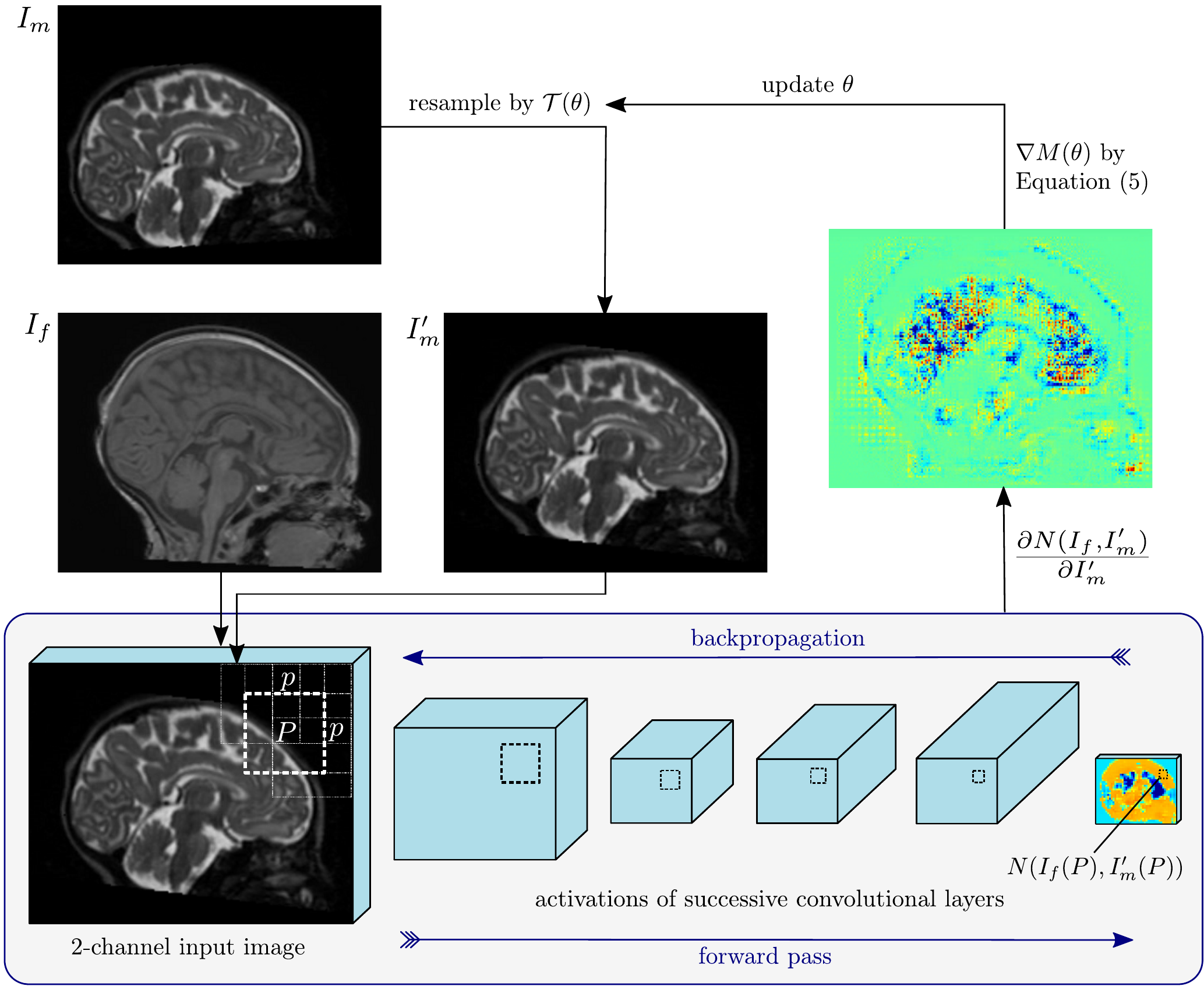}
\caption{\label{fig:overview}
Overview of our method. The input to the network $N$ consists of the fixed image $\fixed$ and the warped image $\warped$ resampled from the moving image $\moving$ by the current transformation $\transf(\param)$. The output of $N$ is a dissimilarity map, where each element corresponds to a patch $P$ in the input (its receptive field). We use its derivative with respect to $\warped$ to update the transformation parameters $\param$. Warm and cold colors corresponds to positive and negative values in the colormaps, respectively.
}
\end{figure*}

\subsection{Related Work}
The idea of using supervised learning to build a similarity metric for multimodal images has  been explored in a number of works. On one side, there are probabilistic approaches which rely on modelling the joint-image distribution. For instance, Guetter \etal propose a generative method based on Kullback-Leibler Divergence~\cite{guetter:kullback}. Our work is closer to the discriminative concept proposed  by Lee \etal \cite{lee:learning} and Michel \etal \cite{michel:boost}, where the problem of learning a similarity metric is posed as binary classification. Here the goal is to discriminate between aligned and misaligned patches given pairs of aligned images. Lee \etal propose the use of a Structured Support Vector Machine while Michel \etal use a method based on Adaboost. Different to these approaches we rely on CNN as our learning method of choice as the suitable set of characteristics for each type of modality combinations can be directly learned from the training data.

The power of CNNs to capture complex relationships between multimodal medical images has been shown in the problem of modality synthesis~\cite{van2015cross}, where  CNNs are used to map MRI-T2 images to MRI-T1 images using jointly the appearance of a small patch together with its localization. Our work is arguably most similar to the approach of Cheng \etal~\cite{cheng:deep} who train a multilayer fully-connected network pretrained with autoencoder for estimating similarity of 2D CT-MR patch pairs. Our network is a CNN, which enables us to scale to 3D due to weight sharing and train from scratch. Moreover, we evaluate our metric on the actual task of registration, unlike Cheng \etal

\section{Method}

Image registration is the task of estimating the best spatial transformation $\transf:\domain_f\mapsto\mathbb{R}^d$ between a \textit{fixed image} $\fixed:\domain_f\subset\mathbb{R}^d\mapsto\mathbb{R}$ and a \textit{moving image} $\moving:\domain_m\subset\mathbb{R}^d\mapsto\mathbb{R}$. In our setting $d=3$ and the images come each from a different modality. The problem is often solved by minimizing the energy 
\begin{equation}
E(\param) = \metric(\fixed,\moving(\transf(\param))) + \regularization(\transf(\param)) \label{eq:Energ}
\end{equation}
where the first term $\metric$ is a metric quantifying the cost of the alignment by transformation $\transf$ parameterized by $\param$ and the second term $\regularization$ is a regularization constraining the mapping. We denote the moving image resampled into $\domain_f$ by $\transf$ as the \textit{warped image} $\warped=\moving(\transf(\param)):\domain_f\subset\mathbb{R}^d\mapsto\mathbb{R}$. The minimization is commonly solved in a continuous or discrete optimization framework~\cite{sotiras:survey}, depending on the nature of $\param$.

In this work we explore formulating $\metric$ as a convolutional neural network. To this end we rely on network  $N(P_f,P_m)$ which outputs a scalar value estimating the dissimilarity between two image patches $P_f\subset\fixed$ and $P_m\subset\warped$ of the same size. Its incorporation into a continuous optimization framework is explained in Subsection~\ref{subsec:co}. The architecture and training of $N$ is described in Subsection~\ref{subsec:nn}.

\subsection{Continuous Optimization} \label{subsec:co}

\def\patchset{{\mathcal{P}}}
\def\xx{{\mathbf{x}}}

Continuous optimization methods iteratively update parameters $\theta$ based on the gradient of the objective function $E(\theta)$. We restrict ourselves to first-order methods and use gradient descent in particular. Our metric is defined to aggregate local patch comparisons as 
\begin{equation}
\metric(\fixed,\warped) = \sum_{P \in \patchset}N(\fixed(P),\warped(P))
\end{equation}
where $\patchset$ is the set of patch domains $P\subset\domain_f$ sampled on a dense uniform grid with significant overlaps. The method is illustrated in Figure~\ref{fig:overview}.

Its gradient $\nabla \metric(\param)$, which is required for $\nabla E(\param)$, can be computed by applying chain rule as follows:
\begin{gather}
\frac{\partial \sum_{P \in \patchset}N(\fixed(P),\warped(P))}{\partial \param} = 
\sum_{\xx \in \domain_f} \sum_{P \in \patchset_{\xx}} {\frac{\partial N(\fixed(P),\warped(P))}{\partial \warped(\xx)} } {\frac{\partial \warped(\xx)}{\partial \param} } = \label{eq:MD1} \\
= \sum_{\xx \in \domain_f} \sum_{P \in \patchset_{\xx}} {\frac{\partial N(\fixed(P),\warped(P))}{\partial \warped(\xx)} } {\frac{\partial \moving(\transf(\param, \xx))}{\partial \transf(\param, \xx)} } {\frac{\partial \transf(\param, \xx)}{\partial \param} } = \label{eq:MD2} \\
= \sum_{\xx \in \domain_f} \frac{\partial N(\fixed,\warped)}{\partial \warped(\xx)} \nabla \moving (\transf(\param, \xx)) J_\transf(\xx) \label{eq:MD3}
\end{gather}

Equation~\eqref{eq:MD1} shows that the derivative of $N$ \wrt the intensity of an input pixel $\xx$ depends on all patches containing it, denoted as $\patchset_{\xx}$. Thus, high overlap of neighboring patches leads to smoother, more stable derivatives. We found that registration quality drops considerably unless the grid stride $s$ of $\patchset$ is small. On the other hand, subsampling $\domain_f$ to obtain a sparser set of samples $\xx$ has a minor impact on performance.

In the transition from Equation~\eqref{eq:MD2}~to~\eqref{eq:MD3}, patch-wise evaluation of $N$ is replaced by fully convolutional evaluation over the whole domain $\domain_f$. This makes the computation very efficient, as results in intermediate network layers can be shared among neighboring patches \cite{Long:fnc}.

Ultimately, the contribution of each pixel $\xx$ to $\nabla \metric(\param)$ is a product of three terms, \cf Equation~\eqref{eq:MD3}: the derivative $\partial N / \partial \warped(\xx)$ of the estimated dissimilarity of patches around $\xx$ \wrt its intensity in the warped image, which can be readily computed by standard backpropagation, the gradient of the moving image $\nabla \moving$, which can be precomputed, and the local Jacobian matrix $J_\transf$ of transformation $\transf$. Note that the choice of a particular transformation type is decoupled from the network, therefore a single network will work with any transformation.

Computing one iteration thus requires resampling of the moving image and one forward and one backward pass in the network. All operations can be efficiently computed on a GPU.

\subsection{Network Architecture and Training}  \label{subsec:nn}

\textbf{Architecture.} A feed-forward convolutional neural network $N$ is used to estimate the dissimilarity of two cubic patches of the same size of $p \times p \times p$ pixels. The architecture is based on recent works on learning to compare patches, notably the 2-channel network of Zagoruyko and Komodakis~\cite{zagoruyko:patches}. The two patches are considered as a 2-channel 3D image (each channel represents a different modality), which is fed to the first layer of the network. The network consists of a series of volumetric convolutional layers with ReLU non-linearities finalized by a convolutional layer without any non-linearity, which produces a scalar score.

To gradually subsample the spatial domain within the network and increase spatial context (perceptive field), we prefer convolutions with non-unit output stride to pooling used in~\cite{zagoruyko:patches}, as it has led to better performance. We hypothesize that too much spatial invariance might be detrimental in our case of learning cross-modal identity, unlike aiming for robustness to distortions such as perspective deformation. The product of convolutional strides determines the overal network stride $s$ used in the fully-convolutional mode.

The 2-channel architecture is powerful as it considers both patches jointly from the beginning. However, its evaluation does not exploit the fact that the fixed image $\fixed$ does not change during optimization and its deep representation could be precomputed in the form of descriptors and cached. We have therefore experimented on  architectures with two independent input branches, such as the pseudo-siamese network in~\cite{zagoruyko:patches}. Unfortunately, we have observed consistent decrease in registration performance.

\textbf{Training.} We suppose to have a set of $k$ aligned pairs of training images $\{(A_j,B_j)\}_{j=1}^k$ with $A_j,B_j: \domain_j\subset\mathbb{R}^d\mapsto\mathbb{R}$. 
We sample transformations $\transf_{i,A_j}$, $\transf_{i,B_j} : \domain_j\mapsto\domain_j$ for $j$-th image pair for data augmentation by varying position, scale, rotation, and mirroring. Patch pairs $X_i=(A_j(\transf_{i,A_j}(P)), B_j(\transf_{i,B_j}(P)))$ with fixed-size domain $P$ are used for training the network. Sample $X_i$ is defined to be positive (labeled $y_i=-1$) if $\transf_{i,A_j}=\transf_{i,B_j}$ and negative ($y_i=1$) otherwise. Positive and negative samples are mined with equal probability. Imposing restrictions on negatives (such as minimum or maximum overlap of source patch domains) or on patch content (such as minimum contrast~\cite{lee:learning}) were experimentally shown detrimental to the registration quality.

The network is trained to classify training samples $X_i$ by minimizing hinge loss $L=\sum_i\max(0, 1-y_i N(X_i))$, which we found to perform better than cross-entropy. We observed that softmax leads to overly flat gradients in continuous optimization, as shown in the bottom plots in Figure~\ref{fig:derivs}. SGD with learning rate 0.01, momentum 0.9 and batch size 128 is used to optimize the network.

Instead of preparing a fixed dataset of patches like in~\cite{cheng:deep}, we sample $X_i$ online. This, together with the augmentations described above, allows us to feed the network with practically unlimited amount of training data. Even for small $k$ we observed no overfitting in learning (see also Subsection~\ref{subsec:trainsz}).

\textbf{Implementation.} We use Torch with cuDNN library for deep learning, elastix  for GPU-based image resampling, and ITK for registration\footnote{www.torch.ch, developer.nvidia.com/cudnn, elastix.isi.uu.nl, www.itk.org}. Our network has 5 layers, 2M parameters, patch size $p=17$, and stride $s=4$. We plan to open source our implementation and the trained network.

\begin{table}[bt]
\centering
\caption{\label{tab:scores}
Overlap scores (mean $\pm$ SD) after registration using the proposed metric (CNN) and mutual information with (MI+M) or without masking (MI)
}
\resizebox{1\linewidth}{!}{
\begin{tabular}{|c|c|c|c|c|c|c|}
\cline{2-7} 
\multicolumn{1}{c|}{} & MI+M & MI & CNN $k=557$ & CNN $k=11$ & CNN $k=6$ & CNN $k=3$\tabularnewline
\hline 
Dice & 0.665 $\pm$ 0.096 & 0.497 $\pm$ 0.180 & 0.703 $\pm$ 0.037 & 0.704 $\pm$ 0.037 & 0.701 $\pm$ 0.040 & 0.675 $\pm$ 0.093\tabularnewline
Jaccard & 0.519 $\pm$ 0.091 & 0.369 $\pm$ 0.151 & 0.555 $\pm$ 0.041 & 0.556 $\pm$ 0.041 & 0.554 $\pm$ 0.044 & 0.527 $\pm$ 0.081\tabularnewline
\hline 
\end{tabular}}
\end{table}

\section{Experiments and Results}

We evaluate the effectiveness of the learned metric in registration experiments on a set of clinical brain images in Subsection~\ref{subsec:mainreg} and conduct further experiments to demonstrate its interesting properties in Subsections~\ref{subsec:trainsz} and~\ref{subsec:derivs}.

\subsection{Deformable Registration of Neonatal Brain MRI Images}  \label{subsec:mainreg}

\textbf{Datasets.} We conducted intersubject deformable registration experiments on a set of neonatal brain image volumes taken from the publicly available brain atlases ALBERTs~\cite{Gousias:alberts1}. This database consists of T1 and T2-weighted MRI scans of 20 newborns. Each T1-T2 pair is aligned and annotated with a segmentation map of 50 anatomical regions, which allows us to evaluate registration quality in terms of overlap measures; we compute average Dice and Jaccard coefficients.

To make the experiment challenging and demonstrate good generalization of our learned metric (denoted CNN), we train on IXI~\cite{ixi:ixi}, a completely independent dataset of adult brain images. Let us remark that there are structural differences between the brains of neonates and adults. The dataset contains about 600 approximately aligned T1-T2 image pairs and we use $k=557$ for training and the rest for validation, although in Subsection~\ref{subsec:trainsz} we demonstrate that much less is actually needed. Image intensities in both datasets are normalized to $[0,1]$.

\textbf{Baseline.} Our baseline is mutual information (MI)~\cite{mattes:mi}, the standard metric for multimodal registration. We observed that MI perform better when image domains are restricted to the head region, thus we use a fixed intensity threshold of $0.01$ for masking the background and denote this variant MI+M. Such masking made nearly no difference to our metric. Unfortunately, we could not compare to other learning-based metrics~\cite{lee:learning,michel:boost} as their implementation was not available.

\textbf{Protocol.} We test on 18 subjects in ALBERTs and perform 68 intersubject registrations, half of them aligning T1 to T2 and half of them the other way round. We reserve the remaining 2 subjects for validating registration parameters and model selection. Both metrics are evaluated in exactly the same registration pipeline with the same transformation model and optimizer. The pipeline consists of multiresolution similarity transform registration followed by multiresolution B-spline registration (2 scales, 1000 control points on the fine scale, 200k image sampling points), optimized by gradient descent with regular step and 500 iterations per scale. MI is used with 75 histogram bins (validated optimum). An explicit regularization term $R$ in Equation~\eqref{eq:Energ} was used neither for MI nor for CNN. Instead, we regularize implicitly by the design of the pipeline and the choice of its hyperparameters.

\begin{figure*}[bt]
\centering
\includegraphics[width=0.36\linewidth]{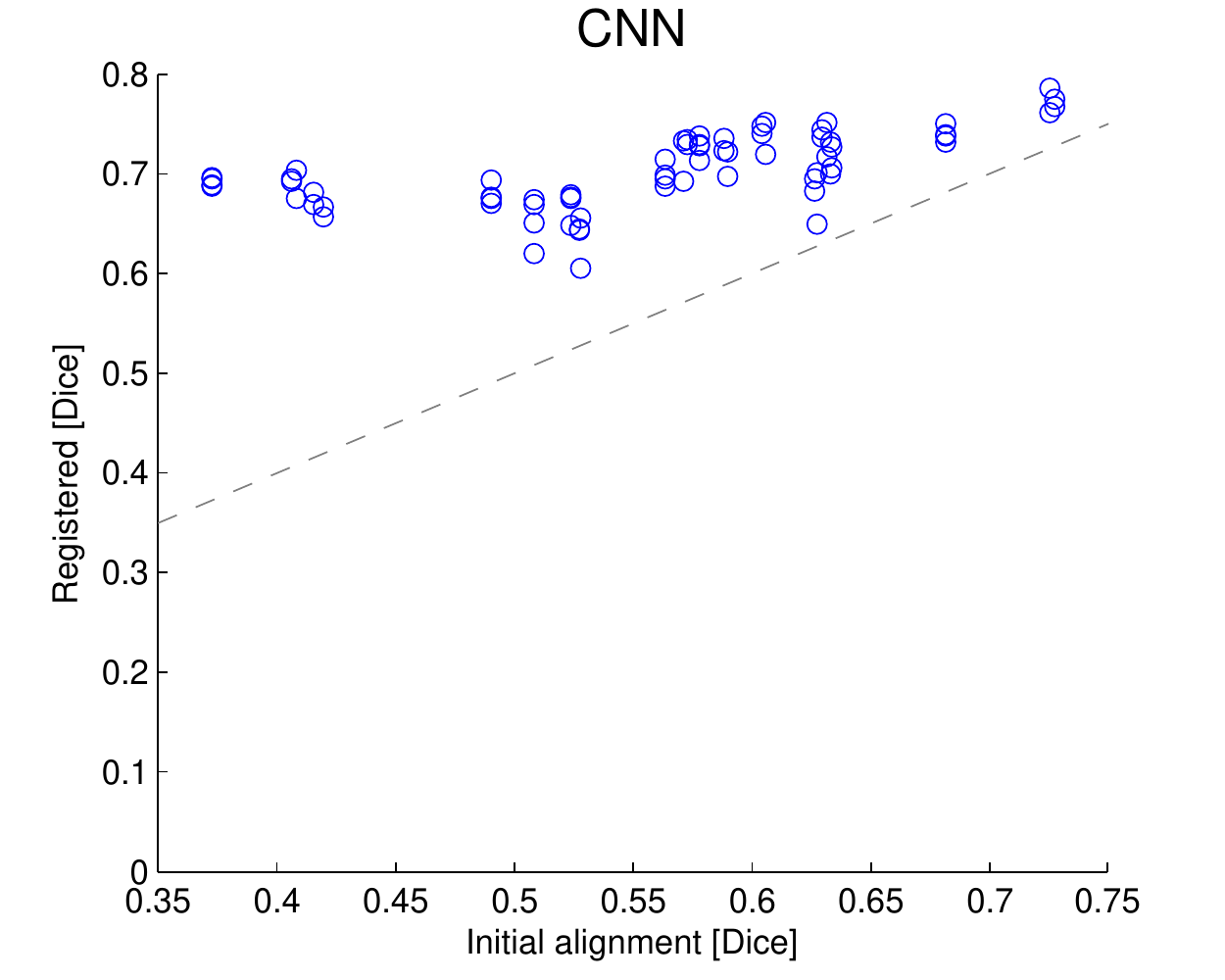}
\hspace{-2em}\includegraphics[width=0.36\linewidth]{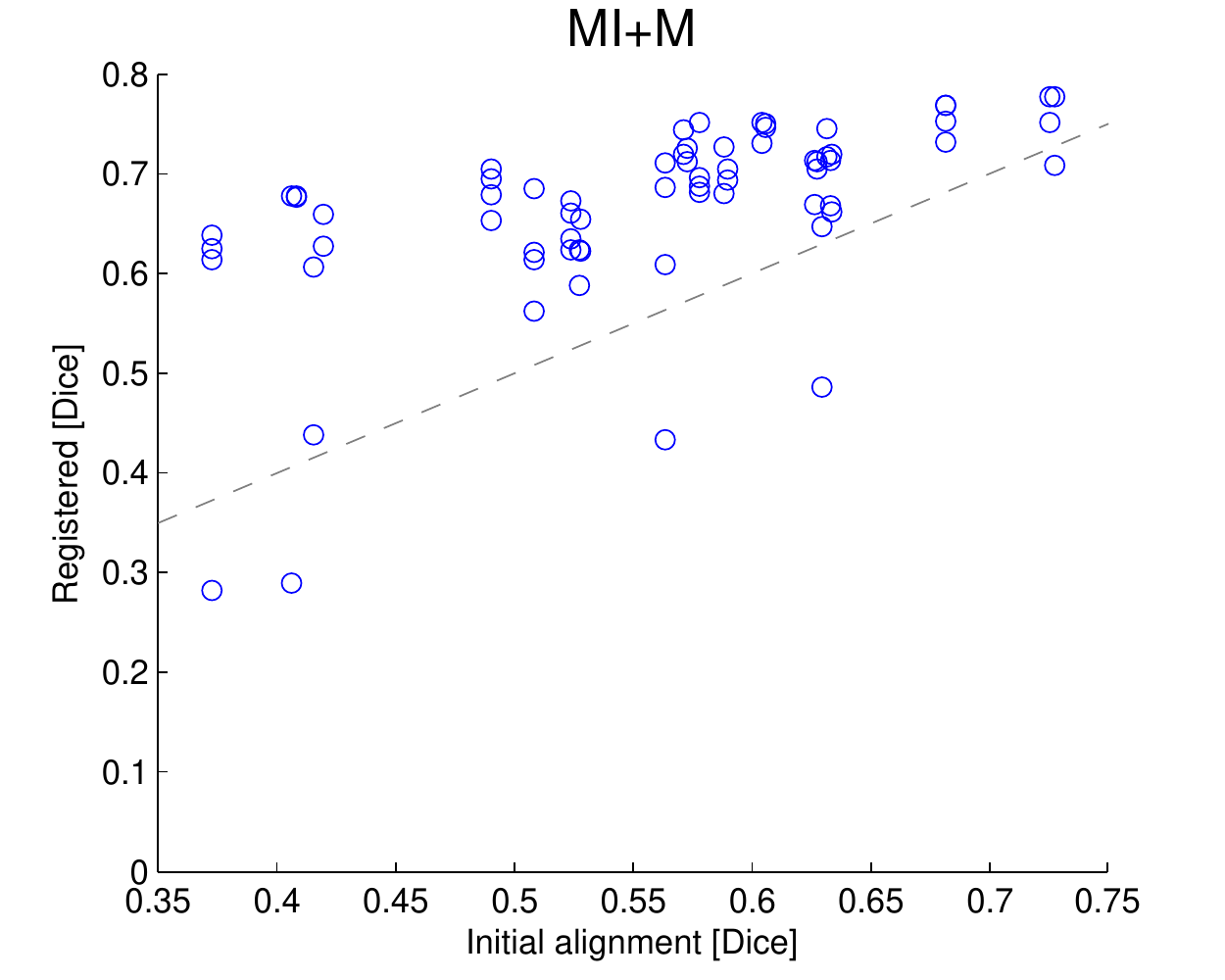}
\hspace{-2em}\includegraphics[width=0.36\linewidth]{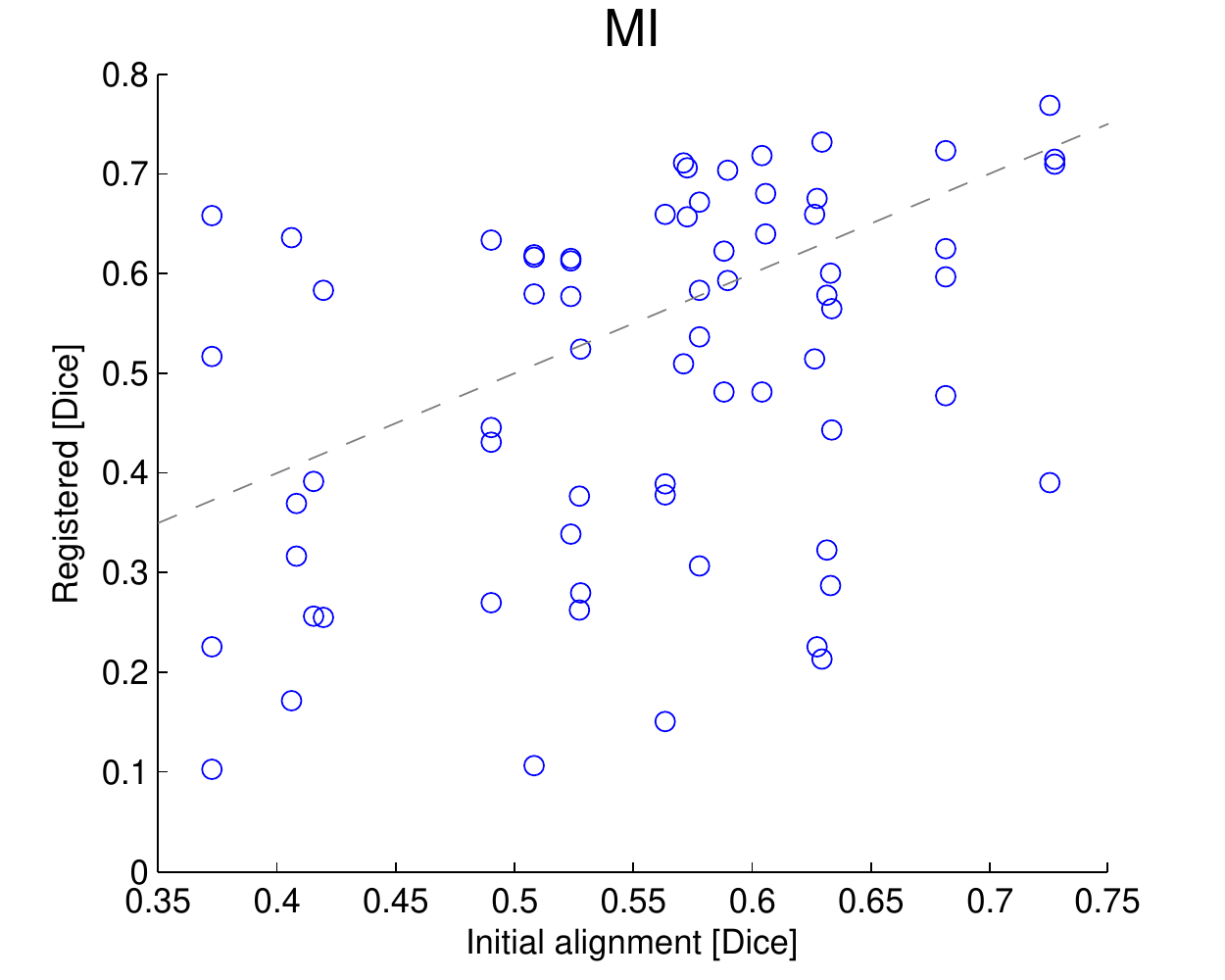}
\caption{\label{fig:scatter}
Improvement in average Dice score due to registration using the proposed metric (CNN) and mutual information with (MI+M) or without masking (MI). Each data point represents a registration run. Dashed line denotes identity transformation.
}
\end{figure*}

\textbf{Results.} The results are listed in Table~\ref{tab:scores} and demonstrate statistically significant improvement of registration quality due to CNN by about 4 points in both coefficients (as by one-sided t-test with significance $\alpha=0.01$). Figure~\ref{fig:scatter} exhibits scatter plots of initial and final Dice scores for each registration run (Jaccard scores follow similar trend). We can see that while CNN has improved on the alignment in all runs, this is not the case for MI+M and especially MI, showing rather low precision. The highest accuracies achieved by both methods are rather similar (up to 0.8) and seem nearly independent on the initial level of misalignment. Furthermore, the registration using CNN is only about 2x slower than using MI (on Nvidia Titan Black), the difference mostly due to expensive resampling of moving image.

\subsection{Influence of Training Set Size}  \label{subsec:trainsz}

The huge number of aligned volumes in IXI dataset is rather exceptional in medical domain. We are therefore interested in how much we can decrease the training set size $k$ without noticeable impact on the quality. To this end, we train networks with only $k=$ 11, 6, and 3 random image pairs under the same setting as above. Table~\ref{tab:scores} shows that even with little training data the results are very good and only for $k=3$ our metric does not significantly outperform MI+M. On one hand, this suggests that our online sampling and data augmentation methodology works well. On the other hand, either the inherent variability in the dataset is very low (especially compared to natural image recognition problems, where more data typically improves performance) or our network is not able to exploit it. We expect that the amount of necessary data will be higher for more challenging modalities, such as ultrasound.

\subsection{Plausibility of Metric and Its Derivatives}  \label{subsec:derivs}

\begin{figure*}[bt]
\centering
\includegraphics[width=0.45\linewidth]{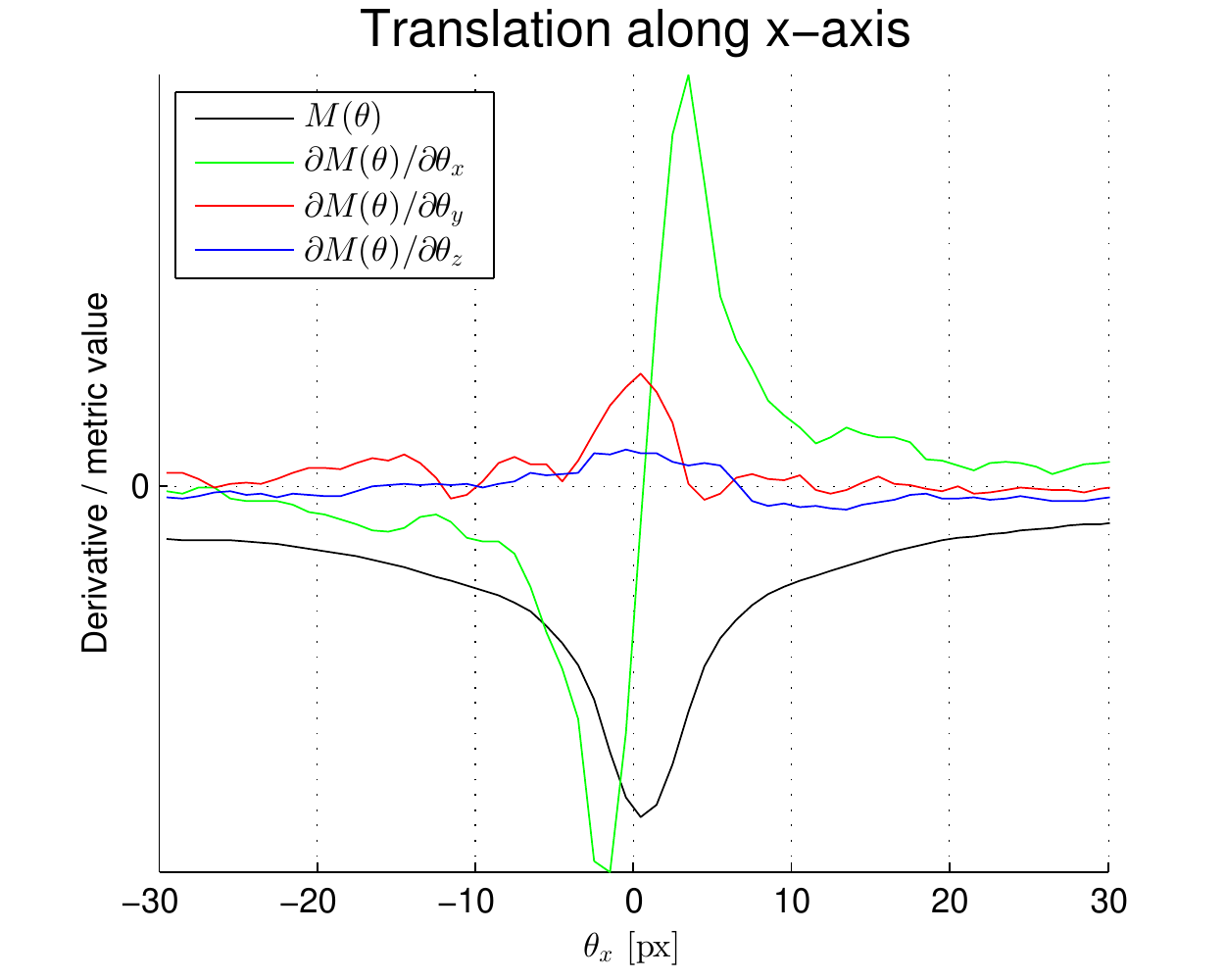}
\includegraphics[width=0.45\linewidth]{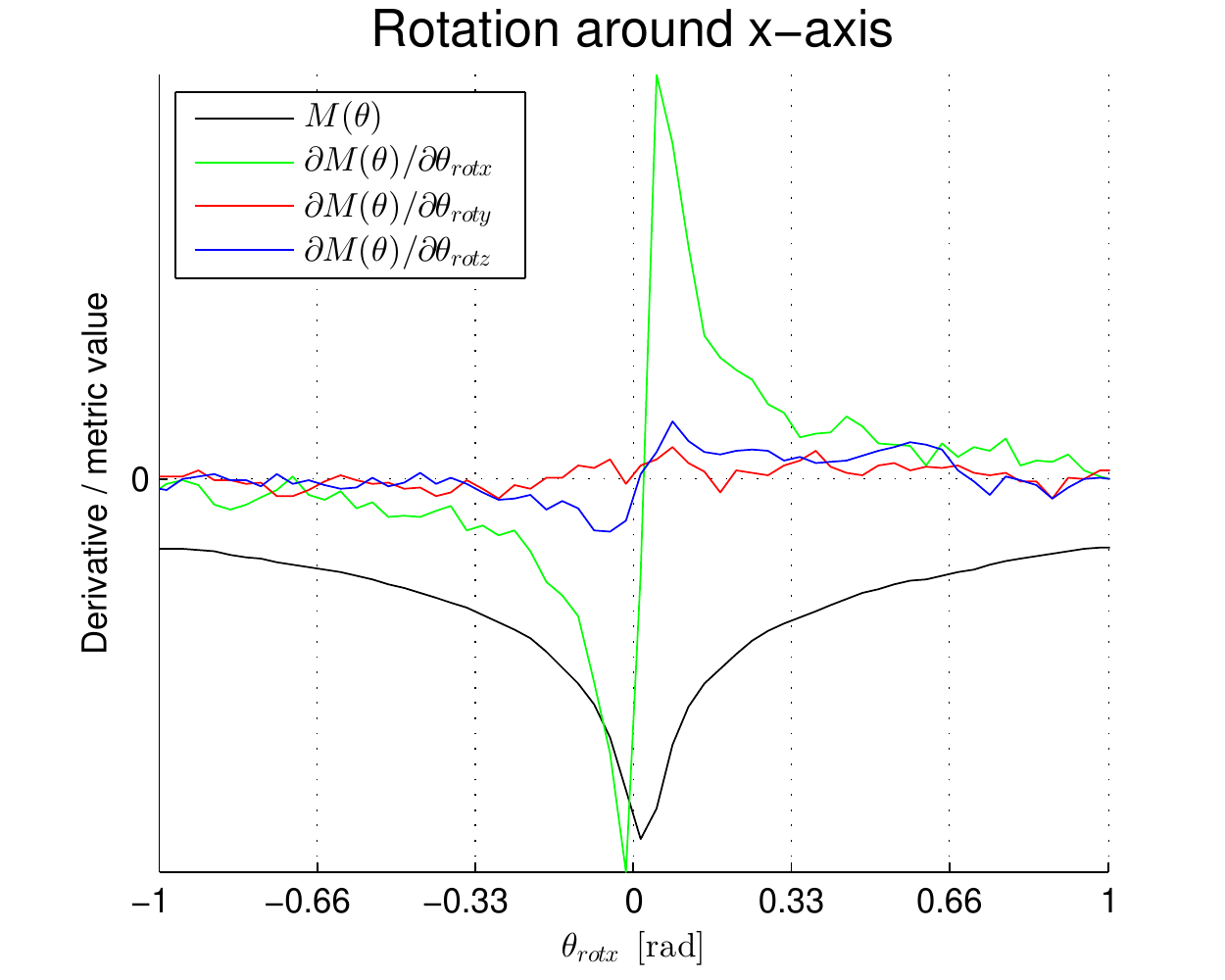}
\includegraphics[width=0.45\linewidth]{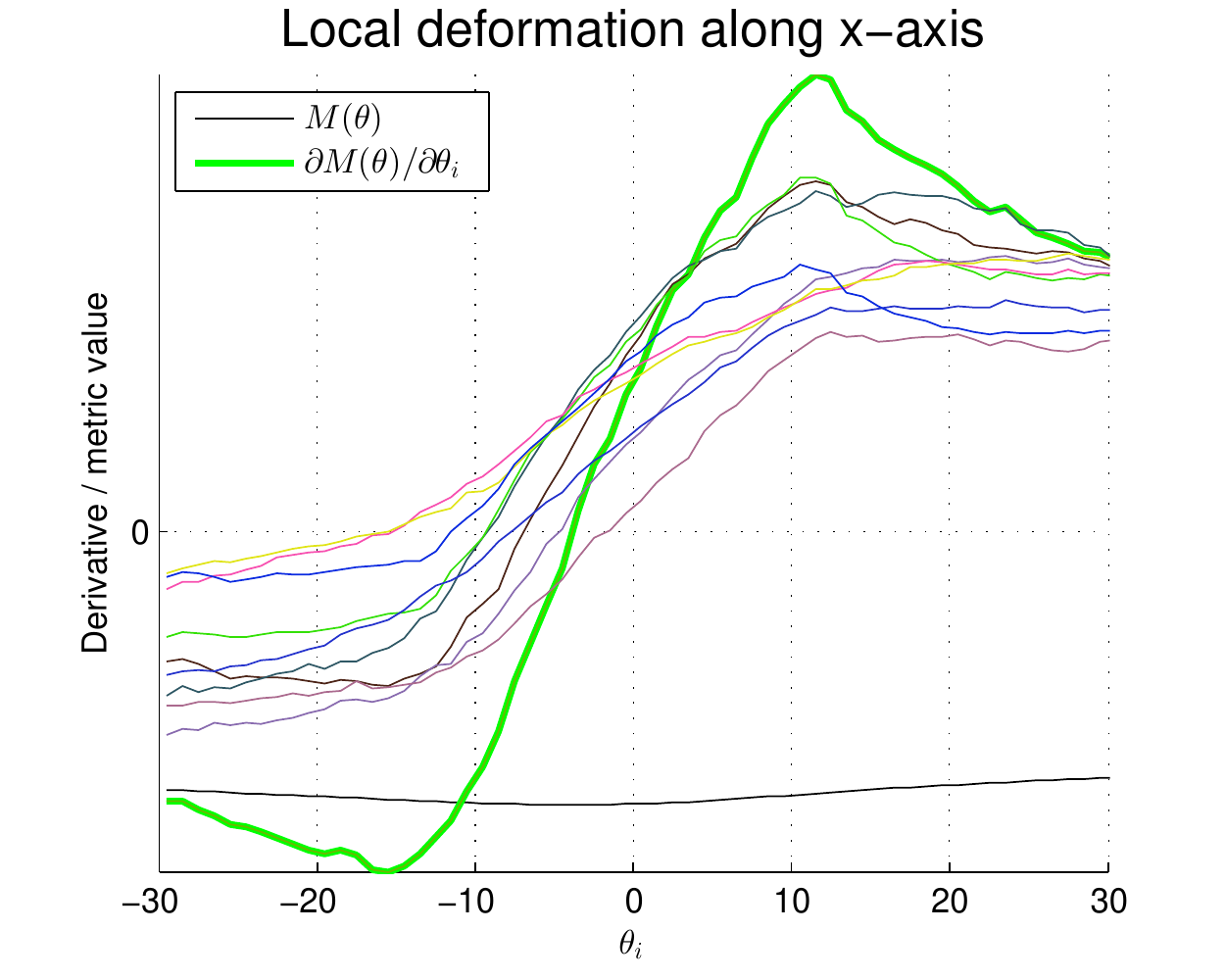}
\includegraphics[width=0.45\linewidth]{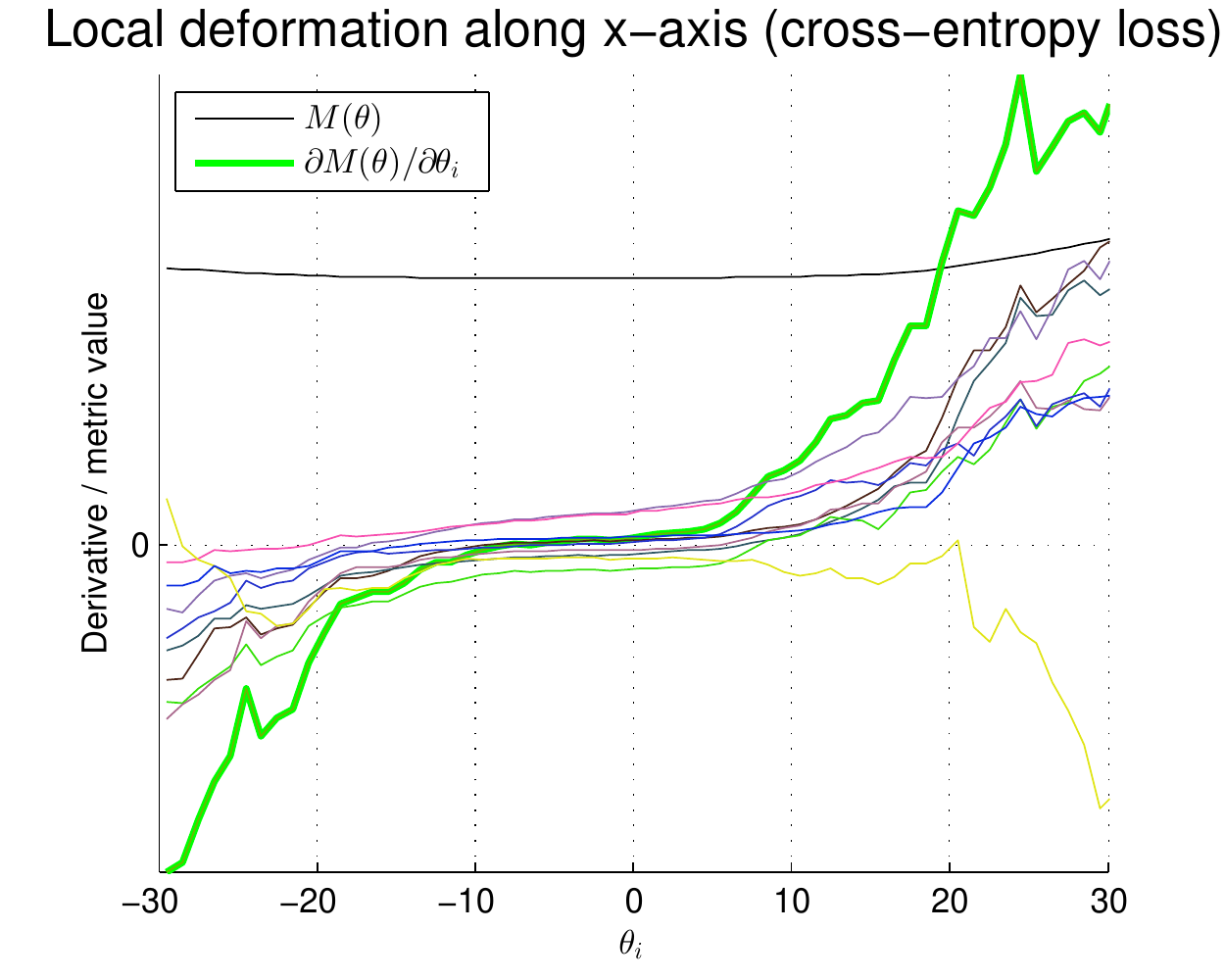}
\caption{\label{fig:derivs}
The impact of perturbation of a single parameter in Euclidean transform (top) and B-spline transform (bottom) on metric value $\metric$ and its derivatives as per Equation~\eqref{eq:MD3}. The curve of $\metric$ is not up to scale. Curves without legend correspond to other parameters strongly affected due to overlapping patches.
}
\end{figure*}

To investigate the behavior of metric value and its actual derivatives used for continuous optimization, we visualize these quantities by manually perturbing a single parameter of a transformation initialized to identity on an aligned validation image pair in IXI. Figure~\ref{fig:derivs} suggests that the metric behaves reasonably as its curves are smooth with the correct local minima. The analytic derivatives, as in Equation~\eqref{eq:MD3}, have the correct sign over a large range, albeit their magnitude is slightly noisy. 
Nevertheless, this was shown not to prevent the metric from obtaining good registration results.

\vspace*{-3ex}
\section{Conclusion}
\vspace*{-1ex}

We have presented a similarity metric for multimodal 3D image registration based on a convolutional neural network. The network can be trained from scratch even from a few aligned image pairs, mostly due to our data sampling scheme. We have described the incorporation of this metric into first-order continuous optimization frameworks. The experimetal evaluation was perfomed on the task of intersubject T1-T2 deformable registration on a dataset different from the one used for training, demonstrating good generalization. In this task, we outperform mutual information by a significant margin.

\def\uu{{\mathbf{u}}}
We envision incorporating our network into a discrete optimization framework as an easy extension. In a MRF-based formulation, the local alignment cost is expressed by unary potentials over nodes in a graph~\cite{glocker:drop}. In particular, a unary potential $g_n(\uu_n)$ related to the cost of assigning a label/translation $\uu_n$ to node $n$ might be defined as $g_n(\uu_n) = N(\fixed(P_n),\moving(\transf(\param,P_n)+\uu_n))$, where $P_n\subset\domain_f$ is a patch domain centered at the control point of transformation $\transf$ corresponding to node $n$. As such an optimization is derivative-free, only the forward pass in the network would be necessary.

We also plan to apply our method to more modalities, such as ultrasound.

\subsubsection*{Acknowledgments.} We gratefully acknowledge NVIDIA Corporation for the donated GPU used in this research. ALBERTs atlases are copyrighted by Imperial College of Science, Technology and Medicine and Ioannis S. Gousias 2013. B. Guti\'errez-Becker thanks the financial support of CONACYT and the DAAD. The official publication is available at Springer via \url{http://dx.doi.org/10.1007/978-3-319-46726-9_2}.

%
%
\bibliographystyle{splncs03}
\bibliography{biblio-macros,biblio}{}

\appendix

\section{Comparison with MIND}

\begin{figure*}[bt]
\centering
\includegraphics[width=0.36\linewidth]{figures/CNN-Dice}
\hspace{-2em}\includegraphics[width=0.36\linewidth]{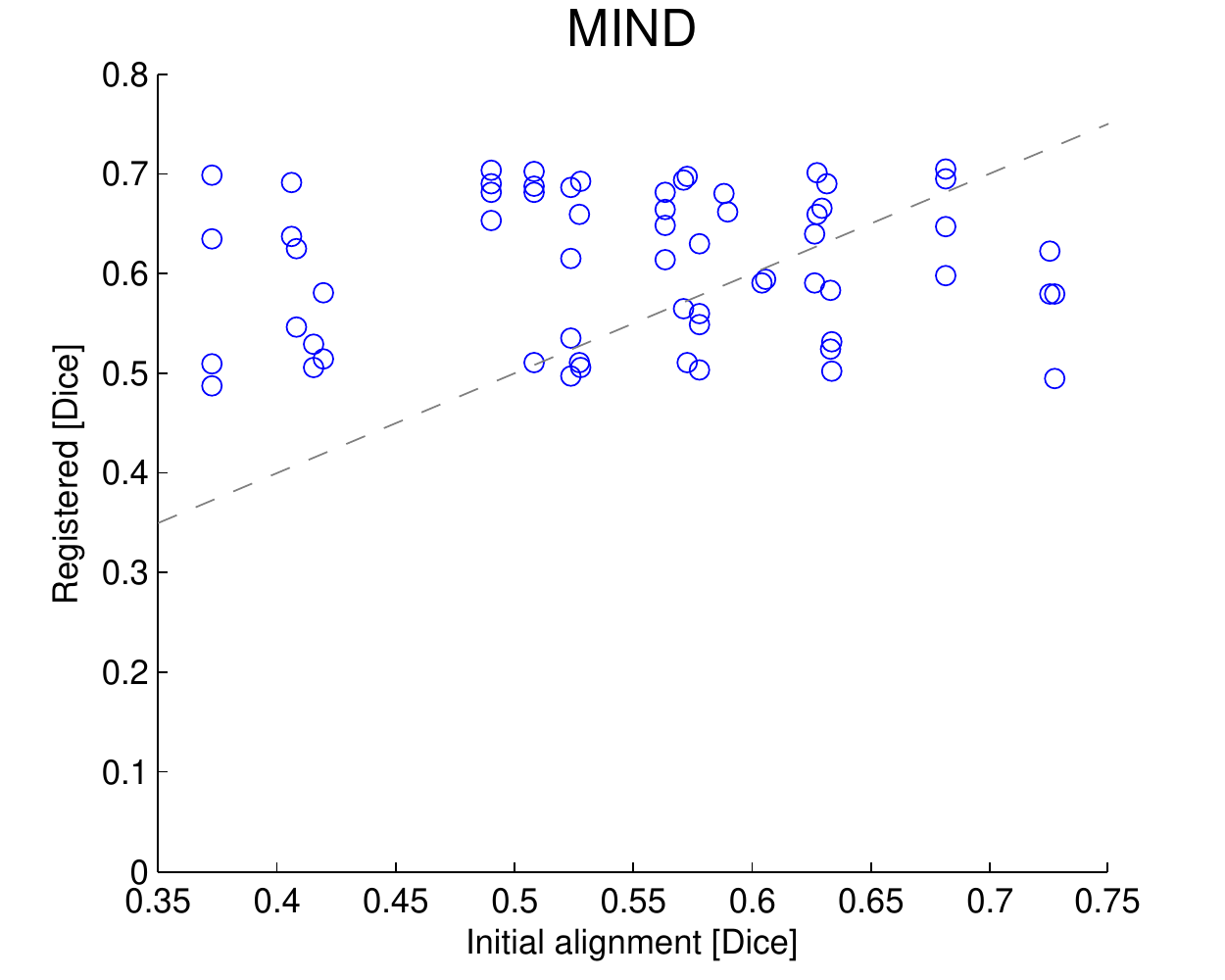}
\caption{\label{fig:scatterMind}
Improvement in average Dice score due to registration using the proposed metric (CNN) and MIND. Each data point represents a registration run. Dashed line denotes identity transformation.
}
\end{figure*}

The Modality Independent Neighborhood Descriptor (MIND) is a state-of-the-art multimodal descriptor by Heinrich \etal~\cite{heinrich:MIND} based on the concept of self-similarity. We performed the same set of 68 registration as in Subsection~\ref{subsec:mainreg} using the code from the authors' website\footnote{http://www.ibme.ox.ac.uk/research/biomedia/julia-schnabel/files/symgn.zip}. After validating its three main hyperparameters (six-neighborhood search region, Gaussian weighting $\sigma=0.5$, regularization $\alpha=0.2$), we obtained Dice score of 0.610 $\pm$ 0.073 (see also Figure~\ref{fig:scatterMind}), resp. Jaccard score of 0.458 $\pm$ 0.070. The results are clearly inferior to both MI+M and our CNN-based approach, although the running time was much shorter. However, we stress that their deformable registration code does not follow our pipeline described in Subsection~\ref{subsec:mainreg} (different regularization and optimization, no similarity transformation step), and therefore the comparison serves for rather illustrative purposes.

\end{document}